\documentclass[conference]{IEEEtran}
\IEEEoverridecommandlockouts

\usepackage{cite}
\usepackage{amsmath,amssymb,amsfonts}
\usepackage{graphicx}
\usepackage{textcomp}
\usepackage{xcolor}
\usepackage{amsmath}
\usepackage{xcolor, bm}
\usepackage{tikz}
\usepackage{pgfplots}
\usetikzlibrary{shapes,arrows}
\usepackage{color, soul}
\usepackage{float}
\usepackage[inline]{enumitem}
\usepackage{multirow}
\usepackage{mathtools}

\usepackage{svg}
\usepackage[a4paper, total={184mm,239mm}]{geometry}
\usepackage{mdframed}
\usepackage{lipsum}
\usepackage{framed}

\usepackage[ruled,linesnumbered]{algorithm2e}

\let\originalparagraph\paragraph
    \renewcommand{\paragraph}[2][.]{\originalparagraph{#2#1}}
    
\begin{document}
\bstctlcite{IEEEexample:BSTcontrol}

\title{\Huge Concurrent Self-testing of Neural Networks Using Uncertainty Fingerprint}
\author{Soyed Tuhin Ahmed, Mehdi B. tahoori\\Department of Computer Science, Karlsruhe Institute of Technology, KIT, Germany}

\maketitle
\addtolength\abovedisplayskip{-0.6em}%
\addtolength\belowdisplayskip{-0.6em}%
\setlength{\textfloatsep}{0pt}
\setlength\belowdisplayskip{0pt}

\begin{abstract}

Neural networks (NNs) are increasingly used in always-on safety-critical applications deployed on hardware accelerators (NN-HAs) employing various memory technologies. Reliable continuous operation of NN is essential for safety-critical applications. During online operation, NNs are susceptible to single and multiple permanent and soft errors due to factors such as radiation, aging, and thermal effects. Explicit NN-HA testing methods cannot detect transient faults during inference, are unsuitable for always-on applications, and require extensive test vector generation and storage.
Therefore, in this paper, we propose the 
\emph{uncertainty fingerprint} approach representing the online fault status of NN. Furthermore, we propose a dual head NN topology specifically designed to produce uncertainty fingerprints and the primary prediction of the NN in \emph{a single shot}. During the online operation, by matching the uncertainty fingerprint, we can concurrently self-test NNs with up to $100\%$ coverage with a low false positive rate while maintaining a similar performance of the primary task. Compared to existing works, memory overhead is reduced by up to $243.7$ MB, multiply and accumulate (MAC) operation is reduced by up to $10000\times$, and false-positive rates are reduced by up to $89\%$.


\end{abstract}


\section{Introduction}

Neural networks (NNs) have become indispensable tools in a multitude of applications ranging from image recognition to natural language processing~\cite{szegedy2016overview, goodfellow2016deep, he2016deep}. Their unprecedented effectiveness has led to their deployment in various sectors, including safety-critical applications such as healthcare, automated industrial predictive maintenance, and autonomous driving~\cite{bojarski2016end, julian2019deep}. NN applications are commonly deployed on hardware accelerator architectures (NN-HA) with different traditional CMOS and emerging memory technologies, such as DRAM~\cite{DrAcc}, SRAM~\cite{9184455}, NAND Flash~\cite{donato2018chip}, 
Resistive RAM (RRAM)~\cite{song2017pipelayer}, Phase Change
Memory (PCM)~\cite{le2018mixed}, Spin-Transfer Torque Magnetic RAM (STT-MRAM)~\cite{pan2018multilevel}. 
In general, NN-HA offers to accelerate the most common NN operation, multiply and accumulated (MAC), to meet the performance and energy consumption needs of high-performance applications including resource-constrained edge applications.

Despite the merits of NN-HA, the memory elements of NN-HA that store weights and activations are susceptible to single and multiple permanent and soft faults that lead to errors~\cite{torres2017fault} due to latent defects from manufacturing test escapes, infant mortality failure mechanisms, random electrical or thermal noise, endurance, aging, and radiation. 
Although NN can tolerate minor perturbations~\cite{ozen2022shaping}, their accuracy decreases significantly with a large-magnitude hardware error~\cite{munch2020tolerating, xia2017fault}, making NN-HA non-functional. 
An incorrect prediction made by a non-functional NN-HA exposed to the end user can lead to catastrophic failures, especially in safety-critical applications. To ensure reliable online operation of NN-HA, a lightweight, yet effective, online concurrent testing approach is required that is suitable for implementation on resource-constrained devices.

Usually, NN-HA receives an input for inference that is sequentially processed by hundreds of layers to produce the prediction. While an NN-HA is performing computations for the prediction, any transient or soft faults cannot be detected by traditional explicit testing methods (pause and test)~\cite{li2019rramedy, Soyed_ITC, luo2019functional}. On the other hand, traditional concurrent error detection (CED), such as error-correcting codes (ECC), are not suitable for detecting multiple, permanent, and logic faults, i.e., faults on NN-HA activations. Also, ECC has a large memory overhead due to parity bit storage.  Furthermore, the error checking (decoding) process has a significant performance and energy penalty~\cite{roberts2014faultsim} and can slow the execution time by $2.5\times$~\cite{gottscho2016measuring}. 
In general, the existing fault detection mechanisms for NN-HA~\cite{li2019rramedy, chen2021line, Soyed_ITC, luo2019functional, meng2021self, meng2022exploring, liu2020monitoring, dos2018analyzing, 9693118, gavarini2022open} either \begin{enumerate*}[label={\alph*)}, font={\color{black!50!black}\bfseries}]
\item infeasible for always-on NN applications as they require stopping the functional operation of NN-HA,
\item requires a large number of test vector generation and storage,
\item needs numerous test queries (forward pass) on the NN-HA for each faulty status check,
\item requires changes to the hardware architecture or access to the internal operation of the NN-HA,
\item can only detect faults in the memory (weights), and
\item involves intensive fault injection studies.
\end{enumerate*}


Therefore, we introduce the \emph{uncertainty fingerprint} approach that fingerprints the status of the offline fault-free NN-HA. By matching the fingerprints during the online operation of the NN-HA, the status of the NN-HA can be checked concurrently (real-time) without interruption. We also propose a dual-head NN topology that is specifically designed to give the \emph{uncertainty fingerprint} as an alternative output parallel to the primary prediction without requiring any extra forward pass.
The effectiveness of the proposed approach is evaluated on multi-bit permanent and soft faults modeled as stuck-at and random flips affecting NN-HA memories (weights) and logic operation for the intermediate results (activation) on various benchmark datasets and topologies. 
The contributions of this paper are summarized as follows:
\begin{enumerate*}
    \item The introduction of the \emph{uncertainty fingerprint} approach along with fingerprint matching-based online concurrent self-testing for NN-HA,
    \item Specifically designed dual-head NN topology for concurrent fault detection while performing the primary task uninterrupted,
    \item Development of a two-stage training strategy, accompanied by an \emph{uncertainty fingerprint} matching loss function, and 
    \item Establishment of a well-defined concurrent testing strategy, along with reduction strategies for false positive alarms. 
\end{enumerate*}


In this paper, we focus on concurrent self-testing of binary NNs (BNNs) as they are more suitable for resource-constrained edge applications but more challenging to self-test. The remainder of the paper is organized as follows. Section~\ref{sec:background} gives a brief background of this work, and reviews existing NN-HA testing techniques. 
Afterward, Section~\ref{sec:proposed} elaborates on different aspects of our proposed approach in detail. 
Later, Section~\ref{sec:results} presents an exhaustive evaluation of our proposed approach; and Section~\ref{sec:conclusion} concludes this paper.

\section{Preliminary}\label{sec:background}

\subsection{Neural Network Topologies}

Convolutional Neural Networks (CNNs), a type of NN, are usually organized as a series of interconnected layers. These layers are typically composed of convolutional, pooling, normalization, and fully connected layers, as well as nonlinear activation functions. In a typical image recognition task, the topology is divided into group feature extraction layers and classifier layers. The feature extraction layers, sometimes also called the backbone network, are responsible for learning representations from the raw input data, while the classifier layers make decisions based on the learned features. 

Traditionally, NN topologies have a single "head," which refers to the last layer responsible for the final output. However, in modern deep learning paradigms, it is increasingly common to have task-specific heads~\cite{zheng2023preventing}. 

\subsection{Defects and Faults in NN Hardware Accelerators}\label{sec:faults}

In NN-HA, defects and faults are largely determined by
its underlying memory technology and architecture. For example, traditional SRAM and DRAM technologies are more susceptible to soft faults caused by alpha particle strikes, cosmic rays, or radiation from radioactive
atoms~\cite{mukherjee2005soft}. However, permanent faults are usually not of great concern ($\sim 12$ failures in 1 billion hours~\cite{ferreira2014extra}) as their manufacturing process is relatively mature. In contrast, emerging memory technologies (RRAM, STT-MRAM, and PCM) are more susceptible to permanent faults due to immature fabrication, imprecise programming,
limited write endurance, and aging. In addition, soft faults can occur due to read disturbances~\cite{bishnoi2014read}, which cause accidental switching of memory cells that share the same bit line during the read operation (inference). Furthermore, due to aging, stored data can change due to retention faults~\cite{munch2020tolerating}. Thus, both permanent and soft faults can affect the weights of the NN-HA. Also, faults can accumulate over time in memory and significantly affect the accuracy of the NN-HA~\cite{munch2020tolerating, xia2017fault}.

In NN-HA, frequently updated intermediate results (MAC results or activation) are typically stored in the buffer memory, e.g. in CMOS latches and flip-flops. 
NN-HA with hybrid memory architectures~\cite{salkhordeh2016operating} uses SRAM/DRAM technologies as buffer memory to store intermediate results, and NVMs are used to store weight values because they are typically less frequently updated~\cite{donato2019memti}. As a result, permanent and soft faults can also occur in NN-HA activations, depending on the memory technologies used to store them. 

We model permanent faults using the "Stuck-at" fault model, where the memory system appears to be held exclusively high or low. This translates to stuck-at-(+1) or stuck-at-(-1) for BNN weights and activation. In the case of multi-bit (K-bit) weights and activation, they can be stuck in any of the $2^k$ states. On the other hand, soft faults are modeled as "random bit flips", implying that the memory element contains random but inaccurate values. For BNN, this translates into $+1$ weights and activations randomly switch to $-1$, and vice versa. However, for multi-bit weights and activations, this means that they can flip to any of the $2^k-1$ other states.

\subsection{Uncertainty Estimation}

Uncertainty in deep learning arises from multiple sources, including, but not limited to, model uncertainty, data uncertainty, and environmental conditions. 
Model uncertainty is inherent due to the approximate nature of NNs, and data uncertainty comes from variations and noise in the training and test datasets. 
The introduction of permanent and soft faults in NN-HA adds another layer of uncertainty to the system. These faults can lead to unpredictable behavior, thereby increasing the overall uncertainty of the NN's outputs. Although challenging, this form of uncertainty is important to quantify, especially in safety-critical applications. In this work, we assume that the inputs that the NN-HA will face in deployment come from a distribution similar to that of the training dataset.

\subsection{Related Works}
Testing methods for NN-HAs can be categorized into several groups such as pause-and-test, self-testing, concurrent testing, and uncertainty estimation methods.

\paragraph{Pause-and-Test Methods} 
These methods, also known as explicit testing,  are some of the most efficient and popular methods to test NNs~\cite{li2019rramedy, chen2021line, Soyed_ITC, luo2019functional}. However, they require a stop in the NN operations for testing, which is impractical for always-on applications. Additionally, they cannot detect most transient faults that occur while NN is computing its prediction. Furthermore, they often require the storage of a large number of test vectors, which can be as much as 10,000~\cite{chen2021line}. Some methods such as~\cite{li2019rramedy} propose the generation of specific test vectors, for example, using adversarial examples, imposing an additional overhead to generate specific test vectors.

\paragraph{Self-Testing Methods}
A couple of studies have proposed to self-test NN~\cite{meng2021self, Soyed_ITC, meng2022exploring, liu2020monitoring} but these methods also require a specific selection of test vectors and storage. Most of the works are efficient, but their drawbacks are similar to the pause-and-test methods. 
Also, some of the image selection process, such as the one proposed by~\cite{meng2021self} involves intensive fault injection studies that have a high computational cost that increases with the size of the model.

\paragraph{Concurrent Test Methods}
Some of the concurrent methods, such as \cite{dos2018analyzing}, require access to the internal operations of the NN. Also, they require an additional checking operation on certain layers (e.g., MaxPool) for every forward pass. Some hardware-based approaches, such as~\cite{9693118}, require changes to the hardware architecture, e.g., implementing adder trees, for constant monitoring, and the dynamic power consumption of the circuit is increased.

\paragraph{Uncertainty Estimation Methods}
In the literature, there are many Bayesian~\cite{gal2016dropout, kingma2013auto} and non-Bayesian~\cite{lakshminarayanan2017simple} methods are proposed to quantify the uncertainty of NN predictions. 
Recently, some works~\cite{jia_efficient_2021, malhotra_exploiting_2020, ahmed_spinbayes_2023, Soyed_Jetcs} focused on implementing them on different accelerators. 
Although uncertainty estimation methods can self-test NNs concurrently, most of these works aim to estimate uncertainty due to data distribution shifts. 
Also, they have a high runtime overhead compared to conventional NN. For example, some methods require multiple forward passes, while other methods require several times the number of parameters.
On the other hand, Open Set Recognition (OSR) methods are low-cost and have recently been studied in work~\cite{gavarini2022open} for detecting single permanent faults affecting NN-HA weights. However, they offer lower fault coverage, higher false positive cases, and are only suitable for image classifications. Also, its adaptability to BNNs, multiple faults, or faults at NN-HA activations was not explored.


\subsection{Problem Statement}\label{sec:problem_statement}

\begin{figure}
    \centering
    \includegraphics[width=0.6\linewidth]{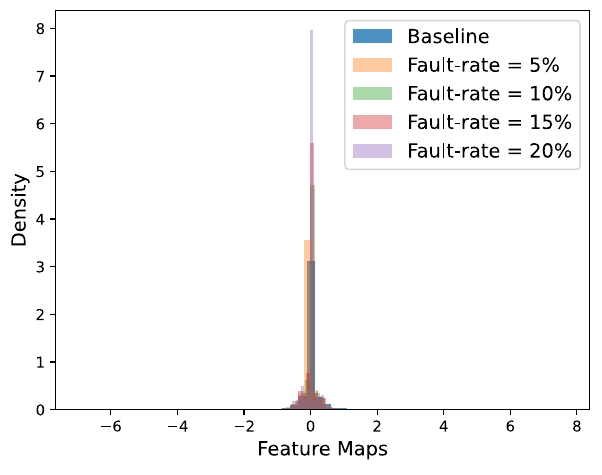}
    \vspace{-1em}
    \caption{Change in the distribution of the feature maps due to soft-faults modeled as bit-flips of weights (see \ref{sec:faults}) on binary ResNet-18 trained on CIFAR-10.}
    \label{fig:CNN_CIFAR10_bitflips_featuremaps}
\end{figure}

As demonstrated in Figure~\ref{fig:CNN_CIFAR10_bitflips_featuremaps}, the learned representations of the backbone network of a ResNet-18 BNN exhibit minimal changes and overlap even at a 20\% soft fault rate. As mentioned previously, these learned representations are received by the classification head, making concurrent self-testing of BNNs a challenging task.

The primary objective is to devise a concurrent self-testing method that meets the following criteria:
\begin{enumerate*}[label={\alph*)}, font={\color{black!50!black}\bfseries}]
    \item Requires only a single forward pass through the network.
    \item Achieves a low False Positive Rate (FPR), minimizing the number of false alarms.
    \item Attains a high True Positive Rate (TPR), ensuring that faults in NN-HA are accurately identified, 
    \item suitable for BNNs, and require lightweight checks online for testing.
\end{enumerate*}


\section{Proposed Method}\label{sec:proposed}

\subsection{Uncertainty Fingerprint}
\begin{figure}
    \centering
    \includegraphics[width=\linewidth]{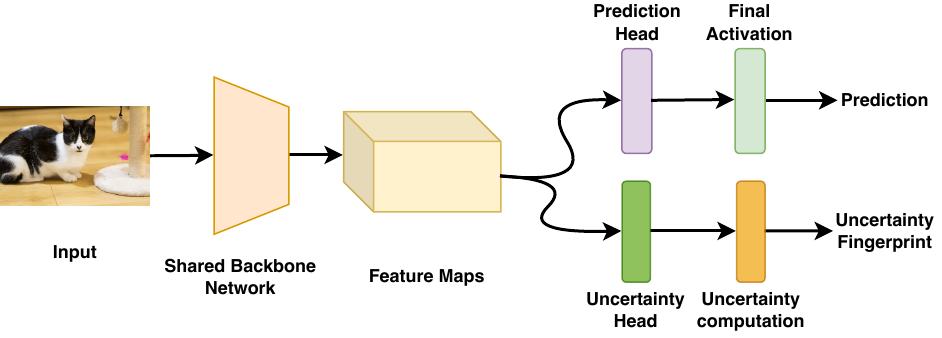}
    \vspace{-1em}
    \caption{Two-Headed model with point estimate parameters and uncertainty for concurrent self-testing. The model is generalizable with existing NN topologies. }
    \label{fig:dual_head}
\end{figure}

In this work, we introduce the \emph{ uncertainty fingerprint} ($\mathcal{F}$) concept, a specialized metric for concurrent fault detection of NN-HAs, especially in always-on safety critical applications. The uncertainty fingerprint is specifically designed to be the output of a dedicated head in a dual-head NN topology as a form of an alternative prediction. We refer to that head as the "uncertainty head." Fig.\ref{fig:dual_head} shows the block diagram of the proposed topology.

In this paper, the uncertainty fingerprint is defined as the maximum value (\texttt{max()}) produced by the uncertainty head. During the training phase, the uncertainty head is explicitly tuned with the objective that the maximum output value approaches \emph{one} for each input. However, the exact value of the uncertainty fingerprint, even in the fault-free state of NN-HA, can vary from one input to another and is itself a distribution. Consequently, optimization makes sure that the uncertainty fingerprint distribution is centered around one. The main goal of the optimization is to establish a "signature" or "fingerprint" based on the NN-HA's fault-free state. Note that the optimization does not require any faulty behavior of NN-HAs or explicit fault injection. 


We hypothesize that as the weights or activation of NN-HA change due to permanent or soft faults, the distribution of uncertainty fingerprint can change, i.e., the distribution shifts to the left or right, thus, making it distinguishable from the pre-defined fault-free distribution. This is because the output of the uncertainty head is a linear transformation of its input, which is the output of the backbone network. As mentioned earlier, faults in the memory elements and intermediate results of NN-HA propagate to the output. That means that both the prediction and the uncertainty head will receive an input that is different from the fault-free one. Consequently, the output of the prediction head will be incorrect, and the uncertainty head will be different from its baseline.

Therefore, by matching the expected uncertainty fingerprint value online for \emph{each prediction}, we propose to concurrently self-test the NN-HA. If the uncertainty fingerprint of the NN-HA matches, then a prediction is classified as fault-free, otherwise, it is classified as faulty. 

\subsection{Dual-Head Model}
As mentioned previously, in modern NN topologies design task-specific heads are increasingly used~\cite{zheng2023preventing}. To obtain the proposed uncertainty fingerprint of the model in a single shot and without reducing the accuracy, we introduce an additional head to the NN, \emph{uncertainty head}. The uncertainty head is typically a linear layer with a predefined number of neurons that is independent of the number of classes in the dataset. The specific number of neurons is a hyperparameter that should be optimized to improve fault detection accuracy.  Unlike conventional NN topologies, our proposed topology can be self-tested. Both the uncertainty and the prediction head share the same unchanged backbone network.

The input received by the uncertainty and prediction head can be the same if the prediction head is also a linear layer. Otherwise, an additional preprocessing layer, such as adaptive average pooling, can be applied before the uncertainty head. The adaptive average pooling layer reduces the spatial dimensions of the feature maps to a single value. Thus, it significantly reduces the size of the weight matrix of the uncertainty head. Hence, our proposed approach can potentially be applied to various NN topologies, such as fully convolutional NN (full-CNN) in which the prediction head is a convolutional layer.


\subsection{Training Objective}\label{sec:training_prop}
We propose a two-step training for our objective. In the first step, the NN is trained using the gradient descent algorithm by minimizing the task-specific loss, e.g., cross-entropy. In this step, the output of the prediction head is only considered.

In the second step, we \emph{freeze} the rest of the model, including the prediction head, and train only the uncertainty head. The term "freeze" here means that the gradient is not calculated with respect to the loss value and the associated parameters and variables will not be updated.

We propose a fingerprint loss function for this step of the training. It is defined as:
\begin{equation}
    \mathcal{L} = \alpha\times
    \frac{1}{N}\sum_{n=1}^{N}(1-\textit{max}(f_{\theta'}(x_n)))^2
\end{equation}

Here, $\alpha$ is a hyperparameter that controls the strength of the loss function, $x_n$ is the NN input,  $f(.)$ denotes the NN with $\theta'$ summarizing all the parameters of the NN excluding the parameters of the prediction head, and $N$ is the batch size. The loss function encourages the uncertainty fingerprint for each input to be close to one. Consequently, it encourages the uncertainty fingerprint distribution to be centered around one. It can be considered as the conventional mean squared error (MSE), but it is applied to the uncertainty head.

For training and baseline uncertainty fingerprint estimation, we divide both the training and validation datasets with an 80:20 split, where 80\% of the training data trains the functional task and 20\% (fingerprint data) trains the uncertainty head to encourage its maximum output close to one. Since the NN was not trained on fingerprint data, the inputs to the uncertainty head from the backbone network resemble those during inference, enabling the uncertainty head to learn outputs akin to those expected during inference, aligning its learned representations closer to the inference scenario. 
Note that random data augmentations and stochastic regularization like Dropout should be avoided during uncertainty head training. Otherwise, the uncertainty head might learn to be robust to input variations, leading to a lower fault detection rate.



\subsection{Online Concurrent Self-test}

We propose boundary-based online testing, which is a lightweight and effective way to detect faults in an online, resource-constrained setting. We pre-compute the boundary values, $l$ and $h$, offline. This range suggests that most of the uncertainty fingerprints of the fault-free model fall within this range. Then, during the online operation, if an uncertainty fingerprint is observed outside this range, it suggests that the data on the memory or intermediate results of NN-HA have changed due to fault occurrence.


The main idea behind this approach is to establish a "normal operating range" for the NN-HA's uncertainty fingerprint. To do this, we compute the $l=2.5\%$ and $h=9.5\%$ quantiles of the uncertainty fingerprint for the fault-free model.
During the online operation, two scenarios can arise regarding the distribution of uncertainty fingerprint score. If the score is less than $l$, it may indicate a leftward drift of the distribution. In contrast, a score greater than $h$ could suggest a rightward drift. In both cases, these shifts signal that the uncertainty fingerprint score is different from what was observed in the fault-free model and the faults in the NN-HA. Therefore, a fault is detected if it satisfies this condition:
\begin{align*}
\text{Status of the model} = 
\begin{cases} 
\text{Faulty} & \text{if } \mathcal{F} < l \text{ or } \mathcal{F} > h, \\
\text{Fault-Free} & \text{otherwise}.
\end{cases}
\end{align*}

As a result, our approach requires only lightweight checks to detect faults. The $<$ (less than) and $>$ (greater than) operations can be implemented in software or even hardware using comparators available in existing hardware accelerators.


Our approach, similar to the work of Gavarini et al.~\cite{gavarini2022open}, may result in false positives or negatives, which need to be minimized. However, unlike~\cite{gavarini2022open}, we propose two strategies to reduce them. The number of false positives or negatives is influenced by the boundaries of the uncertainty fingerprint. Therefore, it is crucial to tune the boundaries for each dataset or task to maximize coverage and minimize false positives.

Initially, we obtain the boundary values from the fault-free model using a dataset that closely represents the expected real-world data distribution that the model will encounter during inference. Thus, the quantiles are determined on $20\%$ of the validation data, providing an unbiased estimate of the uncertainty fingerprint distribution. The remaining $80\%$ of the validation dataset is used for evaluating performance and fault coverage.

Since we choose boundaries as quantile values $2.5\%$ and $95\%$, it effectively ignores the tails of the uncertainty fingerprint distribution, treating them as anomalies. In some datasets and models, these anomalies can skew the values of $l$ and $h$. Therefore, we compute the Z-score of the uncertainty fingerprint distribution and adjust $l$ and $h$ on data with a Z-score of less than two. Z-scores help identify anomalies in a distribution, with scores above two indicating an anomaly.

Note that the boundaries can be adjusted online if necessary, especially when false positives are encountered.
The proposed concurrent detection method serves as an initial line of defense. It can trigger more rigorous (explicit) tests, e.g., to find the location of faults, followed by appropriate mitigation approaches, such as retraining~\cite{yoshikiyo2023nn}.

\section{Experimental Results}\label{sec:results}
\subsection{Simulation Setup}\label{sec:sim_setup}



\subsubsection{Evaluated Models and Dataset}

The proposed method is evaluated across four state-of-the-art CNN topologies: ResNet-18, PreActResNet-34, VGG-9, and SegNet (a full-CNN topology), trained in the CIFAR-10, SVHN, Flowers-102 (classification tasks) and breast cancer segmentation (biomedical semantic segmentation) datasets, respectively. These topologies diverge not only in architectural depth, spanning $11$ to $34$ layers, but also in the number of target classes, ranging from $2$ to $102$. All models have binary ($+1$ and $-1$) weights and activations using the IRNet algorithm~\cite{qin2020forward}, but the bit width of the SegNet activation is increased to 4-bit as the task is much harder.


\subsubsection{Fault-injection}
We perform software-level fault injection to perform Monte Carlo fault simulation with $100$ fault runs. Faults are injected into pre-trained NN weights and activations at random locations given by Bernoulli's distribution. Specifically, bit-flip faults are injected into NN activation during each forward pass to simulate faults occurring while NN-HA computes inference results.


\subsubsection{Evaluation Metrics}
In terms of evaluation metrics, we report fault coverage which is defined as the percentage of validation data flagged as "faulty". In a fault-free model, fault coverage represents FPR and in a faulty model, it represents TRP. In the ideal case, FPR is $0\%$ and TPR is $100\%$. Thus, low FPR and high TPR are desired at the same time. 




\subsection{Inference Accuracy}
\begin{table}[]
\caption{Comparison of the proposed method with the baseline method with different topologies.}
\vspace{-1em}
\centering
\begin{tabular}{|c|cccc|}
\hline
\multirow{2}{*}{Method} & \multicolumn{4}{c|}{Topology}                                                                              \\ \cline{2-5} 
                        & \multicolumn{1}{c|}{ResNet-18} & \multicolumn{1}{c|}{PreActResNet} & \multicolumn{1}{c|}{VGG-9} & SegNet \\ \hline
Baseline                & \multicolumn{1}{c|}{90.68}     & \multicolumn{1}{c|}{95.54}        & \multicolumn{1}{c|}{83.0}    & 93.53  \\ \hline
Proposed                & \multicolumn{1}{c|}{90.68}     & \multicolumn{1}{c|}{95.54}        & \multicolumn{1}{c|}{83.0}    & 93.53  \\ \hline
\end{tabular}
\label{tab:acc}
\end{table}

The proposed dual-head model maintains comparable performance across various NN topologies, as illustrated in Table~\ref{tab:acc}. In particular, both the proposed and baseline models were trained using $80\%$ of the training data (see Section~\ref{sec:training_prop}). A slight improvement in accuracy, estimated between $0.1$ and $1\%$, could be observed if the baseline NNs were trained with $100\%$ of the training dataset.

\subsection{Analysis of Permanent and Soft Faults Coverage}
\begin{figure}
    \centering
    \includegraphics[width=1\linewidth]{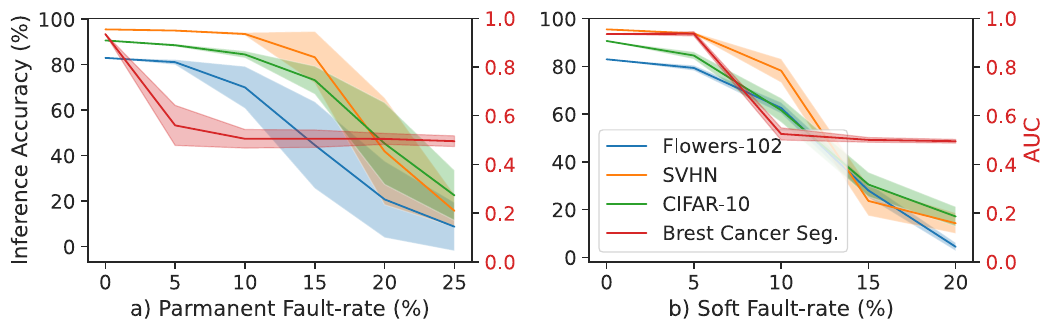}
    \vspace{-2em}
    \caption{Impact of inference accuracy due to (a) permanent faults and (b) soft faults impacting NN-HA weights. Shaded regions indicate the one standard deviation variation around the mean inference accuracy or AUC scores.}
    \label{fig:accs}
\end{figure}

\begin{figure*}
    \centering
    \includegraphics[width=0.9\linewidth]{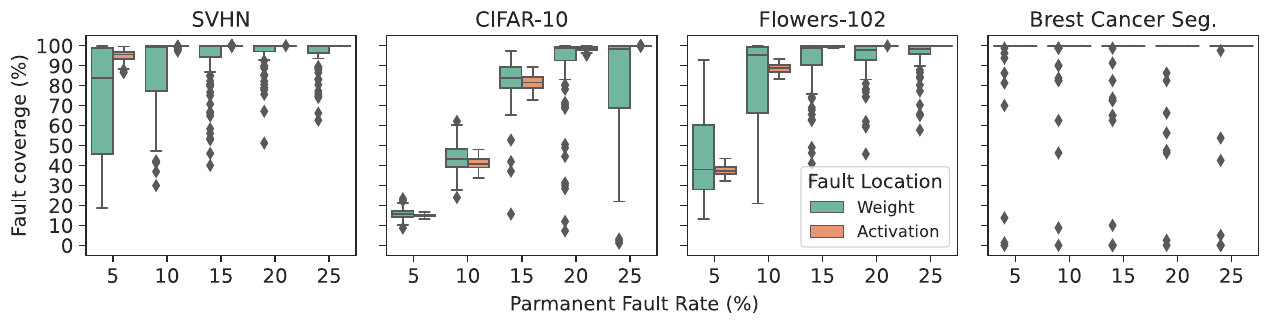}
    \vspace{-1em}
    \caption{Distribution of fault coverage of the proposed method when dealing with permanent faults on weights and activations of NN-HA for various datasets.}
    \label{fig:permanent_faults}
    \vspace{-1em}
\end{figure*}

Figs.~\ref{fig:accs}(a) and (b) show the effect of the permanent and soft faults rate (occurring on the NN-HA weights) on the inference accuracy in different datasets. It can be observed that as the fault rate increases, the inference accuracy decreases across all datasets. According to~\cite{luo2019functional}, an NN-HA is defined as nonfunctional when the inference accuracy drops more than $20\%$ below its baseline accuracy. The classification datasets show a noticeable decline beyond the fault rate of $5\%$ and $10\%$, for permanent and soft faults, respectively. Therefore, at those fault rates, the NN is considered nonfunctional. In the case of the segmentation task, there is a noticeable decrease in the AUC score around the $5\%$ fault rate for both permanent and soft faults. Therefore, from $5\%$ faults, the NN is considered nonfunctional. 
Similar accuracy drop patterns are observed for permanent and soft faults in NN-HA activations.

When the NN-HA is completely nonfunctional due to permanent faults in weights and activations, fault coverage consistently approaches $100\%$ for all datasets, as shown in Fig.~\ref{fig:permanent_faults}. In this scenario, the worst-case median of the fault coverage distribution is $\sim 85\%$ at the $15\%$ fault rate for the CIFAR-10 dataset. Even in a functional case, our proposed method can achieve up to $100\%$ median of the fault coverage distribution.

The change in the uncertainty fingerprint distribution depends on the fault rate and its impact on NN-HA performance. As shown in Fig.~\ref{fig:MLS_Uncer_fingerprint_dist} (left), as the soft fault rate increases, the uncertainty fingerprint distribution moves further away from its baseline.  The amount and direction of drift depend on the dataset, NN topology, fault location, and fault type. Therefore, even at the same fault rate, the fault coverage varies across NN topology, dataset, and fault type. Also, since the accuracy loss at low fault rates is marginal at low fault rates, coverage is also low. For instance, with a $5\%$ permanent fault, the mean accuracy loss on CIFAR-10 is merely $2\%$, resulting in a coverage of around $20\%$ at that fault rate. Given the minimal accuracy drop, such faults are deemed benign, rendering the low coverage relatively harmless.




\begin{figure*}
    \centering
    \includegraphics[width=0.9\linewidth]{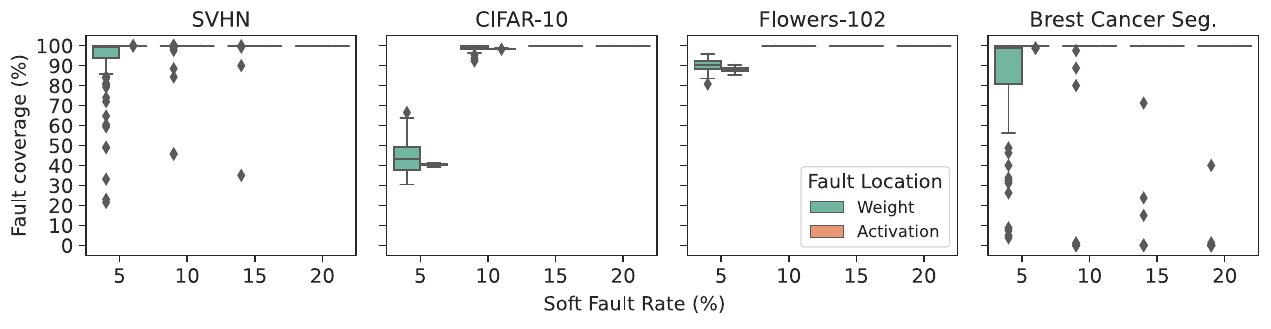}
    \vspace{-1em}
    \caption{Box plots depicting the distribution of fault coverage of the proposed method under soft faults on weights and activations of NN-HA.}
    \label{fig:soft_faults}
    \vspace{-2em}
\end{figure*}


Similarly, in the case of multiple soft faults in NN-HA weights and activations, it can be observed in Fig.~\ref{fig:soft_faults} that the proposed method can consistently achieve $100\%$ fault coverage when NN-HA is nonfunctional. Additionally, fault coverage distributions are significantly more compact compared to permanent faults. Even with a low soft fault rate of $5\%$, most models can achieve near $100\%$ fault coverage. 

\textbf{In summary, on both permanent and soft faults affecting NN-HA weights and activations, the proposed method can consistently achieve a fault coverage distribution with a median of $\mathbf{100\%}$ when the model is non-functional}. Furthermore, relatively high fault coverage can be achieved in the case of marginal degradation in accuracy due to faults. Note that the outliers in the box plots of Figs.~\ref{fig:permanent_faults} and ~\ref{fig:soft_faults} are also rare instances, and fault coverage represents true positive cases. As the true positive rate is consistently high at $100\%$, our method can detect faults accurately with rare false negative instances when the model is not functional.

\subsection{Analysis of FPR and Comparison With Related Works}

Earlier, we showed that the proposed method attains high fault coverage for both permanent and soft faults. However, maintaining low fault coverage in fault-free scenarios is also important to minimize false positive alarms. Specifically, our method exhibits a false positive rate of $11.17\%$, $9.72\%$, $7.25\%$, and $12.5\%$ for the SVHN, CIFAR-10, Flowers-102, and Breast cancer segmentation datasets, respectively.

\begin{figure}
    \centering
    \includegraphics[width=1\linewidth]{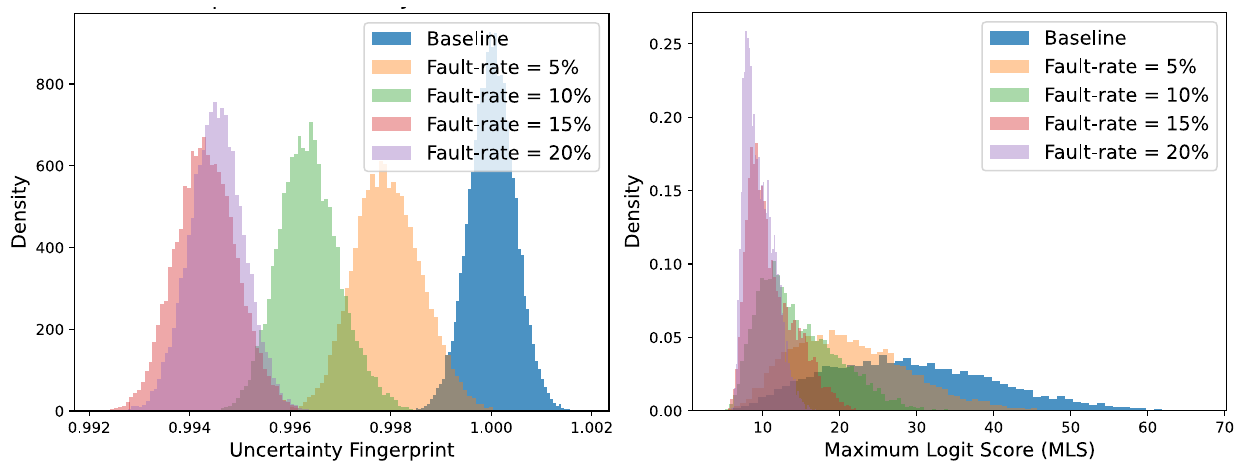}
    \vspace{-2em}
    \caption{Change in the distribution of the proposed uncertainty fingerprint and maximum logit score~\cite{vaze2021open} method due to soft faults on NN-HA weights.}
    \label{fig:MLS_Uncer_fingerprint_dist}
    \vspace{-1em}
\end{figure}

There is a trade-off between the true positive rate and the false positive rate. Calculating the boundaries $l$ and $h$ in the $1.5\%$ and $97\%$ quantiles of the uncertainty fingerprint distribution reduces the false positive rate to $8.65\%$, $6.8\%$, $5.75\%,$ $6.25\%$, but the true positive rate is reduced by $\sim 2\%$. Regardless, our proposed method requires a $62.2\%$ and $89.35\%$ lower false positive rate to achieve a $100\%$ true positive rate compared to OSR-based fault detection~\cite{gavarini2022open}. Specifically, their work requires a false positive rate of $45-69\%$ and $97-98\%$ for CIFAR-10 and SVHN, which is unacceptable for many applications, as it would raise too many false alarms about the state of the model. Also, as shown in Fig.~\ref{fig:MLS_Uncer_fingerprint_dist}, the distribution of the maximum logit score~\cite{vaze2021open}, an OSR method studied in~\cite{gavarini2022open}, overlaps at most of the fault rates, complicating fault detection.

\subsection{Overhead Analysis}


\begin{figure}
    \centering
    \includegraphics[width=\linewidth]{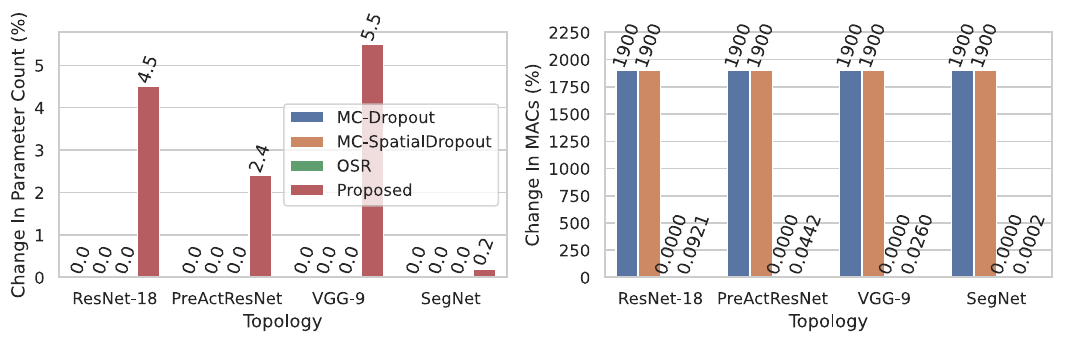}
    \vspace{-2em}
    \caption{Change in the number of parameters and MACs of our method and related works MC-Dropout~\cite{gal2016dropout, Soyed_Jetcs}, MC-SpatialDropout~\cite{rock2021efficient, ahmed2023spatial} respective to the baseline.}
    \label{fig:params_macs}
\end{figure}

\subsubsection{Model Parameters}
Our proposed method increases the number of parameters marginally in all topologies, as shown in Fig.~\ref{fig:params_macs}. For example, the number of parameters in the worst case increases only by $5.5\%$ ($2.05$ MB) for VGG-9, while for SegNet, the increase is as low as $0.2\%$ ($0.26$ MB). Compared to the pause and test methods~\cite{chen2021line, li2019rramedy, luo2019functional, Soyed_ITC}, \textbf{memory overhead to store test vectors is zero for our approach}, but up to $245.775$ MB for existing work, as reported in~\cite{Soyed_ITC}.

\subsubsection{Multiply and Accumulate Operations}
Similarly, our method has only a negligible increase in MACs across all topologies compared to the baseline, as shown in Fig.~\ref{fig:params_macs}. In the worst case, the MACs increase by just $0.0442\%$ over the baseline for the PreActResNet topology. In contrast, MC-Dropout and MC-SpatialDropout methods require $1900\%$ more MACs for each input, assuming they require $20$ forward passes to estimate uncertainty.
Compared to pause and test methods~\cite{chen2021line, li2019rramedy, luo2019functional, Soyed_ITC}, \textbf{our approach does not require additional forward passes}, but they require up to $10000$ forward passes that translate $10000\times$ MACs for each status check of an NN.


\vspace{-0.5em}
\subsection{Discussion and Future Works}
While occasional false positives are not severely harmful, a lack of coverage is. Therefore, the bounds $l$ and $h$ can be adjusted to favor the detection coverage, even if it slightly increases the FPR,
balancing between detection coverage and the false positive rate. In the future, we aim to further improve the fault coverage and reduce FPR by using a contrastive loss function and increasing the depth uncertainty head.

\section{Conclusion}\label{sec:conclusion}
\vspace{-0.2em}

In this paper, we propose a novel approach to the concurrent self-test of NNs using the dual-head model and uncertainty fingerprint. Our approach enables continuous fault monitoring without necessitating the cessation of the primary task, thus fulfilling a critical gap in most of the current research. The proposed dual-head NN topology is specifically designed to produce uncertainty fingerprints and primary predictions in a single shot. During online operation, our approach can concurrently self-test NN-HAs with up to $100\%$ coverage while maintaining the performance of the primary task on benchmark datasets and topologies.

\bibliographystyle{IEEEtran}
\typeout{}
\bibliography{references}

\begin{thebibliography}{10}
\providecommand{\url}[1]{#1}
\csname url@samestyle\endcsname
\providecommand{\newblock}{\relax}
\providecommand{\bibinfo}[2]{#2}
\providecommand{\BIBentrySTDinterwordspacing}{\spaceskip=0pt\relax}
\providecommand{\BIBentryALTinterwordstretchfactor}{4}
\providecommand{\BIBentryALTinterwordspacing}{\spaceskip=\fontdimen2\font plus
\BIBentryALTinterwordstretchfactor\fontdimen3\font minus \fontdimen4\font\relax}
\providecommand{\BIBforeignlanguage}[2]{{%
\expandafter\ifx\csname l@#1\endcsname\relax
\typeout{** WARNING: IEEEtran.bst: No hyphenation pattern has been}%
\typeout{** loaded for the language `#1'. Using the pattern for}%
\typeout{** the default language instead.}%
\else
\language=\csname l@#1\endcsname
\fi
#2}}
\providecommand{\BIBdecl}{\relax}
\BIBdecl

\bibitem{szegedy2016overview}
C.~Szegedy, ``An overview of deep learning,'' \emph{AITP 2016}, p.~7, 2016.

\bibitem{goodfellow2016deep}
I.~Goodfellow, Y.~Bengio, and A.~Courville, \emph{Deep learning}.\hskip 1em plus 0.5em minus 0.4em\relax MIT press, 2016.

\bibitem{he2016deep}
K.~He \emph{et~al.}, ``Deep residual learning for image recognition,'' in \emph{Proceedings of the IEEE conference on computer vision and pattern recognition}, 2016, pp. 770--778.

\bibitem{bojarski2016end}
M.~Bojarski \emph{et~al.}, ``End to end learning for self-driving cars,'' \emph{arXiv preprint arXiv:1604.07316}, 2016.

\bibitem{julian2019deep}
K.~D. Julian, M.~J. Kochenderfer, and M.~P. Owen, ``Deep neural network compression for aircraft collision avoidance systems,'' \emph{Journal of Guidance, Control, and Dynamics}, vol.~42, no.~3, pp. 598--608, 2019.

\bibitem{DrAcc}
\BIBentryALTinterwordspacing
Q.~Deng \emph{et~al.}, ``Dracc: A dram based accelerator for accurate cnn inference,'' in \emph{Proceedings of the 55th Annual Design Automation Conference}, ser. DAC '18.\hskip 1em plus 0.5em minus 0.4em\relax New York, NY, USA: Association for Computing Machinery, 2018. [Online]. Available: \url{https://doi.org/10.1145/3195970.3196029}
\BIBentrySTDinterwordspacing

\bibitem{9184455}
H.~Jiang, R.~Liu, and S.~Yu, ``8t xnor-sram based parallel compute-in-memory for deep neural network accelerator,'' in \emph{2020 IEEE 63rd International Midwest Symposium on Circuits and Systems (MWSCAS)}, 2020, pp. 257--260.

\bibitem{donato2018chip}
M.~Donato \emph{et~al.}, ``On-chip deep neural network storage with multi-level envm,'' in \emph{Proceedings of the 55th Annual Design Automation Conference}, 2018, pp. 1--6.

\bibitem{song2017pipelayer}
L.~Song \emph{et~al.}, ``Pipelayer: A pipelined reram-based accelerator for deep learning,'' in \emph{2017 IEEE international symposium on high performance computer architecture (HPCA)}.\hskip 1em plus 0.5em minus 0.4em\relax IEEE, 2017, pp. 541--552.

\bibitem{le2018mixed}
M.~Le~Gallo \emph{et~al.}, ``Mixed-precision in-memory computing,'' \emph{Nature Electronics}, vol.~1, no.~4, pp. 246--253, 2018.

\bibitem{pan2018multilevel}
Y.~Pan \emph{et~al.}, ``A multilevel cell stt-mram-based computing in-memory accelerator for binary convolutional neural network,'' \emph{IEEE Transactions on Magnetics}, vol.~54, no.~11, pp. 1--5, 2018.

\bibitem{torres2017fault}
C.~Torres-Huitzil and B.~Girau, ``Fault and error tolerance in neural networks: A review,'' \emph{IEEE Access}, vol.~5, pp. 17\,322--17\,341, 2017.

\bibitem{ozen2022shaping}
E.~Ozen and A.~Orailoglu, ``Shaping resilient ai hardware through dnn computational feature exploitation,'' \emph{IEEE Design \& Test}, vol.~40, no.~2, pp. 59--66, 2022.

\bibitem{munch2020tolerating}
C.~M{\"u}nch, R.~Bishnoi, and M.~B. Tahoori, ``Tolerating retention failures in neuromorphic fabric based on emerging resistive memories,'' in \emph{2020 25th Asia and South Pacific Design Automation Conference (ASP-DAC)}.\hskip 1em plus 0.5em minus 0.4em\relax IEEE, 2020, pp. 393--400.

\bibitem{xia2017fault}
L.~Xia \emph{et~al.}, ``Fault-tolerant training with on-line fault detection for rram-based neural computing systems,'' in \emph{Proceedings of the 54th Annual Design Automation Conference 2017}, 2017, pp. 1--6.

\bibitem{li2019rramedy}
W.~Li \emph{et~al.}, ``Rramedy: Protecting reram-based neural network from permanent and soft faults during its lifetime,'' in \emph{2019 IEEE 37th International Conference on Computer Design (ICCD)}.\hskip 1em plus 0.5em minus 0.4em\relax IEEE, 2019, pp. 91--99.

\bibitem{Soyed_ITC}
S.~T. Ahmed and M.~B. Tahoori, ``Compact functional test generation for memristive deep learning implementations using approximate gradient ranking,'' in \emph{2022 IEEE International Test Conference (ITC)}, 2022, pp. 239--248.

\bibitem{luo2019functional}
B.~Luo \emph{et~al.}, ``On functional test generation for deep neural network ips,'' in \emph{2019 Design, Automation \& Test in Europe Conference \& Exhibition (DATE)}.\hskip 1em plus 0.5em minus 0.4em\relax IEEE, 2019, pp. 1010--1015.

\bibitem{roberts2014faultsim}
D.~Roberts and P.~Nair, ``Faultsim: A fast, configurable memory-resilience simulator,'' in \emph{The Memory Forum: In conjunction with ISCA}, vol.~41.\hskip 1em plus 0.5em minus 0.4em\relax Citeseer, 2014.

\bibitem{gottscho2016measuring}
M.~Gottscho \emph{et~al.}, ``Measuring the impact of memory errors on application performance,'' \emph{IEEE Computer Architecture Letters}, vol.~16, no.~1, pp. 51--55, 2016.

\bibitem{chen2021line}
C.-Y. Chen and K.~Chakrabarty, ``On-line functional testing of memristor-mapped deep neural networks using backdoored checksums,'' in \emph{2021 IEEE ITC}, 2021.

\bibitem{meng2021self}
F.~Meng, F.~S. Hosseini, and C.~Yang, ``A self-test framework for detecting fault-induced accuracy drop in neural network accelerators,'' in \emph{Proceedings of the 26th Asia and South Pacific Design Automation Conference}, 2021, pp. 722--727.

\bibitem{meng2022exploring}
F.~Meng and C.~Yang, ``Exploring image selection for self-testing in neural network accelerators,'' in \emph{2022 IEEE Computer Society Annual Symposium on VLSI (ISVLSI)}.\hskip 1em plus 0.5em minus 0.4em\relax IEEE, 2022, pp. 345--350.

\bibitem{liu2020monitoring}
Q.~Liu \emph{et~al.}, ``Monitoring the health of emerging neural network accelerators with cost-effective concurrent test,'' in \emph{2020 57th ACM/IEEE Design Automation Conference (DAC)}.\hskip 1em plus 0.5em minus 0.4em\relax IEEE, 2020, pp. 1--6.

\bibitem{dos2018analyzing}
F.~F. dos Santos \emph{et~al.}, ``Analyzing and increasing the reliability of convolutional neural networks on gpus,'' \emph{IEEE Transactions on Reliability}, vol.~68, no.~2, pp. 663--677, 2018.

\bibitem{9693118}
M.~Liu \emph{et~al.}, ``Online fault detection in reram-based computing systems for inferencing,'' \emph{IEEE Trans. on VLSI Systems}, vol.~30, no.~4, pp. 392--405, 2022.

\bibitem{gavarini2022open}
G.~Gavarini \emph{et~al.}, ``Open-set recognition: an inexpensive strategy to increase dnn reliability,'' in \emph{2022 IEEE 28th International Symposium on On-Line Testing and Robust System Design (IOLTS)}.\hskip 1em plus 0.5em minus 0.4em\relax IEEE, 2022, pp. 1--7.

\bibitem{zheng2023preventing}
Z.~Zheng \emph{et~al.}, ``Preventing zero-shot transfer degradation in continual learning of vision-language models,'' \emph{arXiv preprint arXiv:2303.06628}, 2023.

\bibitem{mukherjee2005soft}
S.~S. Mukherjee, J.~Emer, and S.~K. Reinhardt, ``The soft error problem: An architectural perspective,'' in \emph{11th International Symposium on High-Performance Computer Architecture}.\hskip 1em plus 0.5em minus 0.4em\relax IEEE, 2005, pp. 243--247.

\bibitem{ferreira2014extra}
K.~B. Ferreira \emph{et~al.}, ``Extra bits on sram and dram errors-more data from the field.'' Sandia National Lab.(SNL-NM), Albuquerque, NM (United States), Tech. Rep., 2014.

\bibitem{bishnoi2014read}
R.~Bishnoi \emph{et~al.}, ``Read disturb fault detection in stt-mram,'' in \emph{2014 International Test Conference}.\hskip 1em plus 0.5em minus 0.4em\relax IEEE, 2014, pp. 1--7.

\bibitem{salkhordeh2016operating}
R.~Salkhordeh and H.~Asadi, ``An operating system level data migration scheme in hybrid dram-nvm memory architecture,'' in \emph{2016 Design, Automation \& Test in Europe Conference \& Exhibition (DATE)}.\hskip 1em plus 0.5em minus 0.4em\relax IEEE, 2016, pp. 936--941.

\bibitem{donato2019memti}
M.~Donato \emph{et~al.}, ``Memti: Optimizing on-chip nonvolatile storage for visual multitask inference at the edge,'' \emph{IEEE Micro}, vol.~39, no.~6, pp. 73--81, 2019.

\bibitem{gal2016dropout}
Y.~Gal and Z.~Ghahramani, ``Dropout as a bayesian approximation: Representing model uncertainty in deep learning,'' in \emph{international conference on machine learning}.\hskip 1em plus 0.5em minus 0.4em\relax PMLR, 2016, pp. 1050--1059.

\bibitem{kingma2013auto}
D.~P. Kingma and M.~Welling, ``Auto-encoding variational bayes,'' \emph{arXiv preprint arXiv:1312.6114}, 2013.

\bibitem{lakshminarayanan2017simple}
B.~Lakshminarayanan, A.~Pritzel, and C.~Blundell, ``Simple and scalable predictive uncertainty estimation using deep ensembles,'' \emph{NeurIPS}, 2017.

\bibitem{jia_efficient_2021}
X.~Jia \emph{et~al.}, ``Efficient {Computation} {Reduction} in {Bayesian} {Neural} {Networks} {Through} {Feature} {Decomposition} and {Memorization},'' \emph{IEEE Trans. on Neural Networks and Learning Systems}, vol.~32, Apr. 2021.

\bibitem{malhotra_exploiting_2020}
A.~Malhotra \emph{et~al.}, ``Exploiting {Oxide} {Based} {Resistive} {RAM} {Variability} for {Bayesian} {Neural} {Network} {Hardware} {Design},'' \emph{IEEE Transactions on Nanotechnology}, vol.~19, pp. 328--331, 2020, conference Name: IEEE Transactions on Nanotechnology.

\bibitem{ahmed_spinbayes_2023}
\BIBentryALTinterwordspacing
S.~T. Ahmed \emph{et~al.}, ``{SpinBayes}: {Algorithm}-{Hardware} {Co}-{Design} for {Uncertainty} {Estimation} {Using} {Bayesian} {In}-{Memory} {Approximation} on {Spintronic}-{Based} {Architectures},'' \emph{ACM Transactions on Embedded Computing Systems}, vol.~22, no.~5s, pp. 131:1--131:25, Sep. 2023. [Online]. Available: \url{https://doi.org/10.1145/3609116}
\BIBentrySTDinterwordspacing

\bibitem{Soyed_Jetcs}
------, ``Spindrop: Dropout-based bayesian binary neural networks with spintronic implementation,'' \emph{IEEE Journal on Emerging and Selected Topics in Circuits and Systems}, vol.~13, no.~1, pp. 150--164, 2023.

\bibitem{yoshikiyo2023nn}
S.~Yoshikiyo \emph{et~al.}, ``Nn algorithm aware alternate layer retraining on computation-in-memory for write variation compensation of non-volatile memories at edge ai,'' in \emph{2023 7th IEEE Electron Devices Technology \& Manufacturing Conference (EDTM)}.\hskip 1em plus 0.5em minus 0.4em\relax IEEE, 2023, pp. 1--3.

\bibitem{qin2020forward}
H.~Qin \emph{et~al.}, ``Forward and backward information retention for accurate binary neural networks,'' in \emph{Proceedings of the IEEE/CVF conference on computer vision and pattern recognition}, 2020, pp. 2250--2259.

\bibitem{vaze2021open}
S.~Vaze \emph{et~al.}, ``Open-set recognition: A good closed-set classifier is all you need?'' \emph{arXiv preprint arXiv:2110.06207}, 2021.

\bibitem{rock2021efficient}
J.~Rock \emph{et~al.}, ``On efficient uncertainty estimation for resource-constrained mobile applications,'' \emph{arXiv preprint arXiv:2111.09838}, 2021.

\bibitem{ahmed2023spatial}
S.~T. Ahmed \emph{et~al.}, ``Spatial-spindrop: Spatial dropout-based binary bayesian neural network with spintronics implementation,'' \emph{arXiv preprint arXiv:2306.10185}, 2023.

\end{thebibliography}

\end{document}